\documentclass{article}

\PassOptionsToPackage{numbers, compress}{nips_natbib}

\usepackage[nonatbib, preprint]{nips_2018}

\usepackage[utf8]{inputenc} 
\usepackage[T1]{fontenc}    
\usepackage{hyperref}       
\usepackage{url}            
\usepackage{booktabs}       
\usepackage{amsfonts}       
\usepackage{nicefrac}       
\usepackage{microtype}      
\usepackage[]{caption}

\usepackage{tabularx}
\usepackage{amsmath,amssymb,amsthm}
\usepackage{graphicx}
\usepackage{subcaption}
\usepackage{listings}
\usepackage{upquote}
\usepackage{wrapfig}
\usepackage{lipsum}
\usepackage{array}
\usepackage{multicol}
\usepackage{multirow}
\usepackage{url}
\usepackage{color}
\usepackage{wrapfig,lipsum,booktabs}
\usepackage{makecell}
\usepackage{booktabs}
\usepackage{changepage}
\usepackage[dvipsnames]{xcolor}

\usepackage{algorithm}
\usepackage[noend]{algpseudocode}
\usepackage{tabulary}
\newcolumntype{K}[1]{>{\centering\arraybackslash}p{#1}}

\title{Multi-view Sentence Representation Learning}

\author{
  Shuai Tang \hspace{1cm} Virginia R. de Sa \\
  Department of Cognitive Science, UC San Diego\\
  \texttt{\{shuaitang93,desa\}@ucsd.edu}
}

\begin{document}

\maketitle

\begin{abstract}
Multi-view learning can provide self-supervision when different views are available of the same data. The distributional hypothesis provides another form of useful self-supervision from adjacent sentences which are plentiful in large unlabelled corpora. Motivated by the asymmetry in the two hemispheres of the human brain as well as the observation that different learning architectures tend to emphasise different aspects of sentence meaning, we create a unified multi-view sentence representation learning framework, in which, one view encodes the input sentence with a Recurrent Neural Network (RNN), and the other view encodes it with a simple linear model, and the training objective is to maximise the agreement specified by the adjacent context information between two views.  We show that, after training, the vectors produced from our multi-view training provide improved representations over the single-view training, and the combination of different views gives further representational improvement and demonstrates solid transferability on standard downstream tasks.
\end{abstract}



\section{Introduction}
Multi-view learning methods provide the ability to extract information from different views of the data and enable self-supervised learning of useful features for future prediction  when annotated data is not available \cite{deSa1993LearningCW}. Minimising the disagreement among multiple views helps the model to learn rich feature representations of the data and, also after training, the ensemble of the feature vectors from multiple views can provide an even stronger generalisation ability.

The distributional hypothesis \cite{harris1954distributional} noted that words that occur in similar contexts tend to have similar meaning \cite{Turney2010FromFT}, and the distributional similarity \cite{firth57synopsis} consolidated this idea by stating that the meaning of a word or a sentence can be determined by the company it has. This principle has been widely used in the machine learning community to learn vector representations of human languages. Models built upon distribution similarity don't explicitly require human-annotated training data; the supervision comes from the semantic continuity of the language data, such as text and speech.

Large quantities of annotated data are usually hard to obtain. Our goal is to propose a learning framework built on the ideas of multi-view learning and the distributional hypothesis to learn from unlabelled data. We draw inspiration from the lateralisation and asymmetry in information processing of the two hemispheres of the human brain where, for most adults, sequential processing dominates the left hemisphere, and the right hemisphere has a focus on parallel processing \cite{bryden2012laterality} but both hemispheres have been shown to have roles in literal and non-literal language comprehension \cite{Coulson2005RightHS,Coulson2007ASR}.  

We aim to leverage the functionality of both RNN-based models, which have been widely applied in sentiment analysis tasks \cite{Yang2016HierarchicalAN}, and the linear/log-linear models, which have excelled at capturing attributional similarities of words and sentences \cite{Arora2016ALV,Arora2017ASB,Hill2016LearningDR,Turney2010FromFT} in our multi-view sentence representation learning framework. Our contribution has three folds:


\textbf{\emph{1.}}) A new multi-view sentence representation learning framework is proposed, in which one view is an RNN encoder, and the other is a linear average-on-word-vectors encoder; the agreement between views uses cosine similarity between a pair of sentence representations generated from the two views. 

\textbf{\emph{2.}}) We show that each view gets improved in our efficiently trained multi-view learning framework compared to  single-view training, and the ensemble of two views provides even better results.

\textbf{\emph{3.}}) The proposed model achieves good performance on the unsupervised tasks, and overall outperforms existing unsupervised transfer learning models, and it also shows solid results on supervised tasks, which are either comparable to or better than those of the best unsupervised transfer model.

Our model utilises the distributional similarity as it learns to maximise the agreement among adjacent sentences. Instead of relying on two different input data sources, our model takes the input data, and processes the information in the input data in two independent and distinctive ways, which enables multi-view learning in our proposed learning framework. In addition, our model provides an intriguing hypothesis for a functional role of hemispheric specialisation which emphasises the importance of both hemispheres in language learning and comprehension.
\section{Related Work}

Learning from the context information guided by the distributional similarity has had great success in learning vector representations of words, such as word2vec \cite{Mikolov2013DistributedRO}, GloVe \cite{Pennington2014GloveGV}, and FastText \cite{Bojanowski2017EnrichingWV}. Joint training of word representations and document representations was also proposed in \cite{Le2014DistributedRO}. 

Our learning framework falls into the same category as described in \cite{Kiros2015SkipThoughtV}, which is built on the distributional hypothesis \cite{harris1954distributional} to learn sentence representations. 
Briefly, skip-thought \cite{Kiros2015SkipThoughtV}, as well as FastSent \cite{Hill2016LearningDR}, CNN-LSTM \cite{Gan2017LearningGS}, etc., use an encoder-decoder model, and the training objective is to maximise the likelihood of generating the surrounding sentences given the current sentence as the input to the encoder. The idea is simple, yet its scalability for very large corpora is hindered by the slow decoding process that dominates training time.

A more intuitive approach is to learn a model that maximises the agreement among the representations of adjacent sentences, and minimises agreement among those not adjacent.  A coherence-based learning framework is proposed in \cite{Li2014AMO} that trains a model to classify whether the sentences in a triplet are contiguous in a corpus or not. Additional discourse information is also helpful in learning sentence representations \cite{Jernite2017DiscourseBasedOF,Nie2017DisSentSR} but can be costly when dealing with large corpora.

The closest related work includes Siamese Continuous Bag-of-words (Siamese CBOW) \cite{Kenter2016SiameseCO}, and Quick-thought vectors (QT) \cite{logeswaran2018an}. Both models maximise the agreement between produced vector representations of adjacent sentences; this objective can be trained on an unlabelled corpus efficiently, and produce sentence representations with rich semantics. Siamese CBOW tunes word vectors to increase the cosine similarity of adjacent sentences, while QT optimises two RNNs to encode the current sentence and the sentences
in the context respectively. Although the two RNN encoders are parameterised independently, the way they process the information in the sentences is the same.

The training objective of our model is also to maximise the cosine similarity between adjacent sentences, but, instead of encoding the current sentence and the sentences in the context using the same architecture, our proposed model encodes sentences in two independent and different views. Having two distinctive information processing views encourages the model to encode different aspects of an input sentence, and is beneficial to the future use of the learnt representations.

It is shown in \cite{Hill2016LearningDR} that the consistency between supervised and unsupervised evaluation tasks is much lower than that within either supervised or unsupervised evaluation tasks alone and that a model that performs well on supervised evaluation tasks may fail on unsupervised tasks. \cite{Conneau2017SupervisedLO} subsequently showed that, with a  labelled training corpus, such as SNLI \cite{Bowman2015ALA} and MultiNLI \cite{Williams2017ABC}, the resulting representations of the sentences from the trained model excel in both supervised and unsupervised tasks. Our model is able to achieve good results on both groups of tasks {\bf without} labelled information.

\section{Model Architecture}

Our goal is to marry the \textbf{RNN-based} sentence encoder and the \textbf{avg-on-word-vectors} sentence encoder into a unified learning framework with a simple training objective. It is intuitive to think of training a single model to maximise the agreement between the two views of the same sentence, 
and also between the adjacent sentences based on the distributional similarity \cite{firth57synopsis}. 

The motivation for the idea is that, as mentioned in the prior work, RNN-based encoders process the sentences sequentially, and are able to capture complex syntactic interactions, while the avg-on-word-vectors encoder has been shown to be good at capturing the coarse meaning of a sentence which could be useful for finding paradigmatic parallels \cite{Turney2010FromFT}.

We present a unified learning framework to learn two sentence encoders in two views jointly; after training, the vectors produced from two encoders of the same sentence input are used to compose the sentence representation. The details of our learning framework are described as follows:

\subsection{Encoders in Two Views}
Consider a batch of $N$ contiguous sentences $S=\{s_1,s_2,...,s_N\}$. For  $s_i$ in $S$, there are $M_i$ words, that are transformed into a sequence of word vectors $\mathbf{X}_i = [\mathbf{x}_i^1, \mathbf{x}_i^2, ..., \mathbf{x}_i^{M_i}] \in \mathbb{R}^{300\times M_i}$ using pretrained word vectors, and passed to two encoders $f$ and $g$ to produce two vector representations $\mathbf{z}^f_i$ and $\mathbf{z}^g_i$. The details of the calculation of the representations from $f$ and $g$ are presented in Table \ref{usage}.

\textbf{Encoder $f(\mathbf{X}_i;\mathbf{W}^f)$}: The $f$ encoding function is a bidirectional Gated Recurrent Unit (GRU) \cite{Chung2014EmpiricalEO} that has the dimensionality of $d$ in each direction; it takes a sequence of word vectors and processes them one at a time, then generates a sequence of hidden states $\mathbf{H}_i=[\mathbf{h}_i^1, \mathbf{h}_i^2, ..., \mathbf{h}_i^{M_i}]\in\mathbb{R}^{2d\times M_i}$. The hidden state at the last time step $M_i$ is taken as the representation $\mathbf{z}_i^f=\mathbf{h}_i^{M_i} \in \mathbb{R}^{2d}$. 

\textbf{Encoder $g(\mathbf{X}_i;\mathbf{W}^g)$:} The $g$ encoding function is simply a single-layer feed-forward neural network, which is a trainable linear projection. As found in prior work \cite{Kenter2016SiameseCO,Hill2016LearningDR,Arora2017ASB}, linear/log-linear models perform better on sentence similarity tasks measured by the cosine metric. 
The $\mathbf{z}_i^g$ is calculated as $\mathbf{z}_i^g=\frac{1}{M_i}\sum_{j=1}^{M_i}\mathbf{W}^g\mathbf{x}_i^j
$, where $\mathbf{W}^g \in \mathbb{R}^{2d\times 300}$ is the weight matrix, thus $\mathbf{z}_i^g\in\mathbb{R}^{2d}$.

\subsection{Removing the First Principal Component}

The idea of removing the top principal components as a post-processing step was applied in both \cite{Arora2017ASB, Mu2017AllbuttheTopSA}, and  is incorporated in both the training and testing phases in our learning framework. The \textbf{Power Iteration} \cite{mises1929praktische} method is used to efficiently estimate the top principal component in training, and it is removed by $\mathbf{z}_i=\mathbf{z}_i-\mathbf{u}\mathbf{u}^\top\mathbf{z}_i \label{remove}$. This step is applied on the representations produced from $f$ and $g$ individually, and the details are presented in Section 1 in the supplementary material.



\subsection{Training Objective}


Learning from the distributional similarity is incorporated in our training objective which is to maximise the agreement between the representations of a sentence pair across two views if one sentence in the pair is in the context of the other one. 
The agreement between two views of a sentence pair $(s_i,s_j)$ is defined as $a_{ij}=a_{ji}=\cos(\mathbf{z}_i^f,\mathbf{z}_j^g)+\cos(\mathbf{z}_i^g,\mathbf{z}_j^f)$.
The training objective is to minimise the loss function:
\begin{align}
\mathcal{L}(\mathbf{W}^f,\mathbf{W}^g)=-\sum_{|i-j|\leq c}\log p_{ij}, \text{ \hspace{0.3cm} where \hspace{0.3cm} } p_{ij} =  \frac{e^{a_{ij}/\tau}}{\sum_{n=1}^Ne^{a_{in}/\tau}} \label{loss}
\end{align}
where $\mathbf{W}^f$ contains the parameters in $f$, $\mathbf{W}^g$ is the parameter matrix for $g$, $\tau$ is the trainable temperature term, which is essential for exaggerating the difference between adjacent sentences and those that are not, and the context window $c$, and the batch size $N$ are hyperparameters. 


\begin{table}[t]
    \caption{\textbf{The calculation of representations in training and testing phase.} ``max($\cdot$)'', ``mean($\cdot$)'', and ``min($\cdot$)'' refer to global max-, mean-, and min-pooling over time, which result in a single vector. The table also presents the diversity of the way that a single sentence representation can be calculated.} 
    \fontsize{9}{12}\selectfont
    \centering
    \begin{tabular}{c||c||c|c}
    \midrule
        \multirow{2}{*}{Phase} & \multirow{2}{*}{Training} & \multicolumn{2}{c}{Testing} \\
        \cline{3-4}
         &  & Supervised & Unsupervised  \\
         \hline
        \midrule
        Bi-GRU: $\mathbf{z}^f$ & $\mathbf{h}_i^{M_i}$ & $[\text{max}(\mathbf{H}_i);\text{mean}(\mathbf{H}_i);\text{min}(\mathbf{H}_i);\mathbf{h}_i^{M_i}]$ & $\text{mean}(\mathbf{H}_i)$\\
        \midrule
        Linear: $\mathbf{z}^g$ & $\text{mean}(\mathbf{W}^g\mathbf{X}_i)$ & $[\text{max}(\mathbf{W}^g\mathbf{X}_i);\text{mean}(\mathbf{W}^g\mathbf{X}_i);\text{min}(\mathbf{W}^g\mathbf{X}_i)]$ & $\text{mean}(\mathbf{W}^g\mathbf{X}_i)$\\
        \midrule
        Ensemble & & Concatenation & Addition \\
        \midrule
    \end{tabular}
    \label{usage}
\end{table}

\section{Experimental Design}
3 unlabelled corpora from different genres are used in our experiments, including the BookCorpus (\textbf{C1}) \cite{Zhu2015AligningBA}, the UMBC News Corpus (\textbf{C2}) \cite{han2013umbc_ebiquity} and the Amazon Book Review (\textbf{C3}) \cite{McAuley2015ImageBasedRO}; the models are trained separately on each of the three corpora. The summary statistics of the three corpora can be found in Table \ref{stats}. The Adam optimiser \cite{Kingma2014AdamAM} and gradient clipping \cite{Pascanu2013OnTD} are applied for stable training. 

\begin{table}[h]
\fontsize{9}{12}\selectfont
\caption{\textbf{Summary statistics} of the three corpora used in our experiments. For simplicity, the three corpora will be referred to as C1, C2 and C3 in the following tables respectively.}
\begin{center}
\begin{tabular}{|c|c|c|c|}
\hline
    Name & \# of sentences & mean \# of words per sentence\\
    \hline
    BookCorpus (\textbf{C1}) & ~74M & 13 \\
    UMBC News (\textbf{C2}) & ~134.5M & 25 \\
    Amazon Book Review (\textbf{C3}) & ~150.8M & 19 \\
    \hline
\end{tabular}
\end{center}
\label{stats}
\end{table}

All of our experiments including training and testing are done in PyTorch\footnote{http://pytorch.org/}. The modified SentEval\footnote{https://github.com/facebookresearch/SentEval/} package with the step that removes the first principal component is used to evaluate our models on the downstream tasks. The hyperparameters $N$, $d$ and $c$ are tuned only on the averaged performance on STS14 of the model trained on the BookCorpus; STS14/C1 performance is thus marked with a $\star$ in Table \ref{unsupervised} and Table \ref{FS_QT} to indicate possible overfitting on that dataset/model only.

\begin{table}[t]
\fontsize{8.5}{12}\selectfont
\caption{\textbf{Results on unsupervised evaluation tasks} (Pearson's $r\times 100$) .
\textbf{Bold} numbers are the best results among unsupervised transfer models, and \underline{underlined} numbers are the best ones among all models. Our models outperform other unsupervised transfer models, and provide comparable results with supervised transfer models. The model trained on UMBC news corpus gives the best results among our three models.}
\begin{center}
\begin{tabular}{| K{0.8cm} || K{0.45cm} | K{0.4cm} | K{0.4cm} | K{0.3cm} K{0.3cm} K{0.3cm} K{0.5cm} | K{0.3cm} K{0.4cm} | K{0.4cm} | K{0.5cm} || K{0.6cm} | K{0.7cm} | K{0.7cm} |}
\hline
\multirow{3}{*}{Task} & \multicolumn{11}{c||}{Un. Transfer \cite{Arora2017ASB,Mu2017AllbuttheTopSA,Kenter2016SiameseCO}} & \multicolumn{3}{c|}{Su. Transfer \cite{Conneau2017SupervisedLO,Wieting2017RevisitingRN}} \\
\cline{2-15}
 & \multicolumn{3}{c|}{\textbf{Multi-view}} & \multicolumn{4}{c|}{GloVe} & \multicolumn{2}{c|}{word2vec} & \multirow{2}{*}{ST}  & S- & Infer & \multirow{2}{*}{GRAN} & LSTM \\
\cline{2-10}
 & C1 & C2 & C3 & avg & tfidf & WR & proc. & bow & proc. &  & cbow & Sent &  & avg \\
 \hline
STS12 & 60.9 & \textbf{64.0} & 60.7 & 52.5 & 58.7 & 56.2 & 54.1 & 57.2 & 57.7 & 30.8 & 47.5 & 58.2 & 62.5 & \underline{64.8} \\
STS13 & 60.1 & \textbf{61.7} & 59.9 & 42.3 & 52.1 & 56.6 & 57.7 & 56.8 & 58.0 & 24.8 & 42.9 & 48.5 & \underline{63.4} & 63.1 \\
STS14 & 71.5$^\star$ & \textbf{73.7} & 70.7 & 54.2 & 63.8 & 68.5 & 59.2 & 62.9 & 63.3 & 31.4 & 60.4 & 67.1 & \underline{75.9} & 75.8 \\
STS15 & 76.4 & \underline{\textbf{77.2}} & 76.5 & 52.7 & 60.6 & 71.7 & 57.3 & 62.7 & 63.4 & 31.0 & 30.7 & 71.1 & 75.8 & 76.7 \\
STS16 & 75.8 & \underline{\textbf{76.7}} & 74.8 & - & - & - & - & - & - & - & - & 71.2 & - & - \\
SICK14 & 74.7 & \underline{\textbf{74.9}} & 72.8 & 69.4 & 69.4 & 72.2& 67.9 & 70.1 & 61.5 & 49.8 & - & 73.4 & 72.9 & 71.3 \\
\hline
\end{tabular}
\end{center}
\label{unsupervised}
\end{table}

\subsection{Unsupervised Evaluation - Textual Similarity Tasks}
\textbf{Representation:} For a given sentence input $s$ with $M$ words, suggested by \cite{Pennington2014GloveGV, Levy2015ImprovingDS}, the representation is calculated as $\mathbf{z}=\hat{\mathbf{z}}^f+\hat{\mathbf{z}}^g$, where $\hat{\mathbf{z}}$ refers to the post-processed and normalised vector, and is mentioned in Table \ref{usage}.

\textbf{Tasks}: The unsupervised tasks include five tasks from SemEval Semantic Textual Similarity (STS) in 2012-2016 \cite{Agirre2015SemEval2015T2,Agirre2014SemEval2014T1,Agirre2016SemEval2016T1,Agirre2012SemEval2012T6,Agirre2013SEM2S} and the SemEval2014 Semantic Relatedness task (SICK-R) \cite{Marelli2014ASC}. 

\textbf{Comparison}: We compare our models with 
(I.) \emph{Unsupervised transfer learning}: Skip-thought (ST), avg-GloVe, tfidf-GloVe, GloVe+WR, GloVe+proc. \cite{Arora2017ASB}, word2vec+bow, word2vec+proc. \cite{Mu2017AllbuttheTopSA}, Siamese CBOW \cite{Kenter2016SiameseCO}, FastSent \cite{Hill2016LearningDR}, and QT \cite{logeswaran2018an}. 
(II.) \emph{Supervised transfer learning}: The avg-LSTM and the GRAN \cite{Wieting2017RevisitingRN} trained on the Paraphrase Database (PPDB) \cite{Ganitkevitch2013PPDBTP}, the InferSent\footnote{We download the released InferSent \cite{Conneau2017SupervisedLO}, and evaluated the model using the modified SentEval package.} \cite{Conneau2017SupervisedLO} trained on SNLI \cite{Bowman2015ALA} and MultiNLI \cite{Williams2017ABC}. The results are presented in Table \ref{unsupervised}. Since the performance of FastSent and QT was only evaluated on STS14, we compare to their results in Table \ref{FS_QT}.

All three models trained with our learning framework outperform other unsupervised transfer learning methods, and the model trained on the UMBC New Corpus gives the best performance among our three models. We found that STS tasks contain multiple news- and headlines-related datasets, and UMBC News Corpus matches the domain of these datasets, which might explain the good results provided the model trained with UMBC News. The detailed results on all datasets in STS tasks are presented in the supplementary material. 

\begin{table}[t]
\caption{Comparison with FastSent and QT on STS14 (Pearson's $r\times 100$).}
\fontsize{8.5}{12}\selectfont
\begin{center}
\begin{tabular}{|c|c|c|c|c|c|c|c|}
\hline
\multicolumn{2}{|c|}{FastSent \cite{Hill2016LearningDR}} & \multicolumn{2}{c|}{QT \cite{logeswaran2018an}} & \multicolumn{3}{c|}{\textbf{Multi-view}} \\
\hline
& +AE & RNN & BOW & C1 & C2 & C3 \\
\hline
61.2 & 59.5 & 49.0 & 65.0 & 71.5$^\star$ & \textbf{73.7} & 70.7 \\
\hline
\end{tabular}
\end{center}
\label{FS_QT}
\end{table}



\subsection{Supervised Evaluation}
The evaluation on these tasks involves learning a linear model on top of the learnt sentence representations produced by the model, thus it is named supervised evaluation. Since a linear model is capable of picking the most relevant dimensions in the feature vectors to make predictions, it is preferred to concatenate various types of representations to a richer and possibly, more redundant feature vector, which allows the linear model to explore the combination of different aspects to provide better results.


\textbf{Representation:}  Inspired by \cite{McCann2017LearnedIT}, the representation $\mathbf{z}^f$ is calculated by concatenating the outputs from the global mean-, max- and min-pooling on top of the hidden states $\mathbf{H}$, and the last hidden state, and $\mathbf{z}^g$ is calculated with three pooling functions as well. The post-processing and the normalisation step is applied individually. These two representations are concatenated  to form a final sentence representation. Table \ref{usage} presents the calculation of the representations.

\textbf{Tasks}: Semantic relatedness (SICK) \cite{Marelli2014ASC}, paraphrase detection (MRPC) \cite{Dolan2004UnsupervisedCO}, question-type classification (TREC) \cite{Li2002LearningQC}, movie review sentiment (MR) \cite{Pang2005SeeingSE}, Stanford Sentiment Treebank (SST) \cite{Socher2013RecursiveDM}, customer product reviews (CR) \cite{Hu2004MiningAS}, subjectivity/objectivity classification (SUBJ) \cite{Pang2004ASE}, opinion polarity (MPQA) \cite{Wiebe2005AnnotatingEO}. 

\textbf{Comparison}: Our results as well as related  results of supervised task-dependent training models, supervised transfer learning models, and unsupervised transfer learning models are presented in Table \ref{supervised}. Note that, for the fair comparison, we collect the results of the best single model (MC-QT) trained on BookCorpus in \cite{logeswaran2018an}. 

The three models trained with our learning framework either outperform other existing methods, or achieve similar results on some tasks. The model trained on the Amazon Book Review gives the best performance on sentiment analysis tasks, since the corpus conveys strong sentiment information.

\begin{table}[t]
\fontsize{8.5}{12}\selectfont
\caption{\textbf{Results on the supervised evaluation tasks.} 
\textbf{Bold} numbers are the best results among unsupervised transfer models, and \underline{underlined} numbers are the best ones among all models. ``\dag'' refers to an ensemble of two models. ``\ddag'' indicates that additional labelled discourse information is required. Our models perform similarly or better than existing methods, but with higher training efficiency.}
\begin{center}
\begin{tabular} {|l| K{0.4cm} | K{1.0cm} K{1.0cm} K{1.0cm} | K{0.6cm} K{0.5cm} K{0.5cm} K{0.5cm} K{0.7cm} K{0.6cm}|}
\hline
\Xhline{1.2pt}
Model & Hrs & SICK-R & SICK-E & MRPC & TREC & MR & CR & SUBJ & MPQA & SST \\
\Xhline{1.2pt}
\multicolumn{11}{c}{\emph{Supervised task-dependent training - No transfer learning}} \\
\hline
AdaSent \cite{Zhao2015SelfAdaptiveHS} & - & - & - & - & 92.4 & 83.1 & \underline{86.3} & 95.5 & \underline{93.3} & - \\
TF-KLD \cite{Conneau2017SupervisedLO} & - & - & - & \underline{80.4}/\underline{85.9} & - & - & - & - & - & - \\
\hline
\multicolumn{11}{c}{\emph{Supervised training - Transfer learning}} \\
\hline
InferSent  \cite{Conneau2017SupervisedLO} & $<$24 & \underline{88.40} & \underline{86.3} & 76.2/83.1 & 88.2 & 81.1 & 86.3 & 92.4 & 90.2 & 84.6 \\
\hline
\multicolumn{11}{c}{\emph{Unsupervised training with unordered sentences}} \\
\hline
TF-IDF \cite{Hill2016LearningDR} & - & - & - & 73.6/81.7 &   85.0   & 73.7 & 79.2 & 90.3 & 82.4 & - \\
ParagraphVec \cite{Le2014DistributedRO} & 4 & - & -& 72.9/81.1 & 59.4 & 60.2 & 66.9 & 76.3 & 70.7 & - \\
word2vec+bow \cite{Conneau2017SupervisedLO} & 2 & 80.30 & 78.7 & 72.5/81.4 & 83.6 & 77.7 &   79.8   & 90.9 &   88.3   &  79.7 \\
GloVe+bow \cite{Conneau2017SupervisedLO} & - & 80.00 & 78.6 & 72.1/80.9 & 83.6 &   78.7   & 78.5 &   91.6   & 87.6 & 79.8 \\
GloVe+WR \cite{Arora2017ASB} & - & 86.03 & 84.6 & - / - & - & - & - & - & - & 82.2 \\
FastText+bow \cite{Mikolov2017AdvancesIP} & - & - & - & 73.4/81.6 & 84.0 & 78.2 & 81.1 & 92.5 & 87.8 & 82.0  \\
SDAE \cite{Hill2016LearningDR} & 72 & - & - & 73.7/80.7 & 78.4 & 74.6 & 78.0 & 90.8 & 86.9 & -  \\
\hline
\multicolumn{11}{c}{\emph{Unsupervised training with ordered sentences}} \\
\hline
FastSent \cite{Hill2016LearningDR} & 2 & - & - & 72.2/80.3 & 76.8 & 70.8 & 78.4 & 88.7 & 80.6 & - \\
FastSent+AE \cite{Hill2016LearningDR} & 2 & - & - & 71.2/79.1 & 80.4 & 71.8 & 76.5 & 88.8 & 81.5 & - \\
ST \cite{Kiros2015SkipThoughtV} & 336 & 85.80 & 82.3 & 73.0/82.0& 92.2 & 76.5 & 80.1 & 93.6 & 87.1 & 82.0 \\
ST+LN \cite{Ba2016LayerN} & 720 & 85.80 & 79.5 & - / - & 88.4 & 79.4 & 83.1 & 93.7 & 89.3 & 82.9 \\
CNN-LSTM \cite{Gan2017LearningGS} \dag & - & 86.18 & - & 76.5/83.8 & 92.6 & 77.8 & 82.1 & 93.6 & 89.4 & - \\
\hline
DiscSent \cite{Jernite2017DiscourseBasedOF} \ddag & 8 & - & - & 75.0/ - & 87.2 & - & - & 93.0 & - & - \\
DisSent \cite{Nie2017DisSentSR} \ddag & - & 79.10 & 80.3 & - / - & 84.6 & 82.5 & 80.2 & 92.4 & 89.6 & 82.9 \\
MC-QT \cite{logeswaran2018an} & 11 & 86.80 & - & 76.9/\textbf{84.0} & \underline{\textbf{92.8}} & 80.4 & 85.2 & 93.9 & 89.4 & -\\
\hline
\textbf{Multi-view} C1 & 3 & \textbf{87.85} & 84.8 & \textbf{77.1}/83.4 & 91.8 & 81.6 & 83.9 & 94.5 & 89.1 & 85.8 \\
\textbf{Multi-view} C2 & 8.5 & 87.82 & \textbf{85.2} & 76.8/83.9 & 91.6 & 81.5 & 82.9 & 94.7 & 89.3 & 84.9 \\
\textbf{Multi-view} C3 & 8 & 87.74 & \textbf{85.2} & 75.7/82.5 & 89.8 & \underline{\textbf{85.0}} & \textbf{85.7} & \underline{\textbf{95.7}} & \textbf{90.0} & \underline{\textbf{89.6}} \\
\hline
\end{tabular}
\end{center}
\label{supervised}
\end{table}

\section{Discussion}
\subsection{Multi-view Learning vs. Single-view Learning}

In order to determine if the multi-view learning with two different views/encoding functions is helping the learning, we compare our model to
other reasonable variants, including the multi-view model with two functions of the same type but parameterised independently, either two $f$-s 
or two $g$-s, and the single-view model with only one $f$ or $g$. The comparison is conducted on both BookCorpus and UMBC news. Table \ref{UMBCmvsv} presents the results of the models trained on UMBC Corpus.\footnote{The comparison on BookCorpus can be found in the supplementary material.}

\textbf{In our multi-view learning with $f$ and $g$, the two encoding functions augment each other view.} As illustrated in previous work, and specifically emphasised in \cite{Hill2016LearningDR}, linear/log-linear models, which include $g$ in our model, produce better representations for STS tasks than RNN-based models do. The same finding can be observed in Table \ref{UMBCmvsv} as well, where $g$ consistently provides better results on STS tasks than $f$ does. In addition, as we expected, in our multi-view learning with $f$ and $g$, $g$ augments the performance of $f$ on STS tasks. With maximising the agreement between the representations generated from $f$ and $g$, it is also expected that $f$ improves $g$ on other supervised evaluation tasks, 
as shown in the table.

\begin{table}[t]
\caption{\textbf{Results of our multi-view model with $f$ and $g$ and other variants}. In the table, ``Avg of STS tasks'' refers to the mean Pearson's score on five STS tasks; ``Avg of SICK-R, STS-B'' refers to the mean Pearson's score on Sick-Entailment and STS-Benchmark as they both require the same feature engineering methods proposed in \cite{Tai2015ImprovedSR}; ``Avg of Binary-CLS tasks'' refers to the mean accuracy on five sentiment analysis tasks; $en(\cdot,\cdot)$ stands for an ensemble of two representations. The arrow indicates the performance boost (\textcolor{ForestGreen}{$\uparrow$}) or drop (\textcolor{red}{$\downarrow$}) relative to the same part in our model, e.g., \textcolor{red}{$\downarrow$17.7} indicates the performance of $\mathbf{z}^f$ in multi-view with $f_1$ and $f_2$ is 17.7 point lower than that of the $\mathbf{z}^f$ in our multi-view model with $f$ and $g$. Better view in colour. }
\fontsize{9}{12}\selectfont
\begin{center}
    \begin{tabular}{ c | c | c | c | c | c }
    \hline 
        \midrule
        \textbf{UMBC News} & \multirow{2}{*}{Hrs} & Avg of STS tasks  & Avg of & Avg of Binary-CLS tasks & \multirow{2}{*}{MRPC} \\[0.5ex]
         (C2) & & (STS12-16) & SICK-R, STS-B & (MR, CR, SUBJ, MPQA, SST) &  \\[0.5ex]
        \Xhline{2pt}
        \midrule
        \multicolumn{6}{c}{\textbf{Our Multi-View with $f$ and $g$}: $a_{ij}=\cos(\mathbf{z}_i^f,\mathbf{z}_j^g)+\cos(\mathbf{z}_i^g,\mathbf{z}_j^f)$} \\[0.5ex]
        \midrule
        $\mathbf{z}^f$ & \multirow{3}{*}{8} & 67.4 & 83.0 & 86.6 & 75.5/82.7\\
        $\mathbf{z}^g$ & & 69.2 & 82.6 & 85.2 & 74.3/82.7 \\
        $en(\mathbf{z}^f, \mathbf{z}^g)$ & & 70.6 & 83.0 & 86.6 & 76.8/83.9 \\
        \Xhline{2pt}
        \midrule
        \multicolumn{6}{c}{Multi-view with $f_1$ and $f_2$: $a_{ij}=\cos(\mathbf{z}_i^{f_1},\mathbf{z}_j^{f_2})+\cos(\mathbf{z}_i^{f_2},\mathbf{z}_j^{f_1})$} \\[0.5ex]
        \multicolumn{6}{c}{Multi-view with $g_1$ and $g_2$: $a_{ij}=\cos(\mathbf{z}_i^{g_1},\mathbf{z}_j^{g_2})+\cos(\mathbf{z}_i^{g_2},\mathbf{z}_j^{g_1})$} \\[0.5ex]
        \midrule
        $\mathbf{z}^{f_1}$ & \multirow{2}{*}{17} & 49.7 \textcolor{red}{($\downarrow$17.7)} & 82.2 \textcolor{red}{($\downarrow$0.8)} & 86.3 \textcolor{red}{($\downarrow$0.3)} & 75.9/83.0  \\
        $en(\mathbf{z}^{f_1}, \mathbf{z}^{f_2})$ & & 57.3 \textcolor{red}{($\downarrow$13.3)} & 81.9 \textcolor{red}{($\downarrow$1.1)} & 87.1 \textcolor{ForestGreen}{($\uparrow$0.5)} & 77.2/83.7 \\
        \midrule
        $\mathbf{z}^{g_1}$ & \multirow{2}{*}{2} & 68.5 \textcolor{red}{($\downarrow$0.7)} & 80.8 \textcolor{red}{($\downarrow$1.8)} & 84.2 \textcolor{red}{($\downarrow$1.0)} & 72.5/82.0 \\
        $en(\mathbf{z}^{g_1}, \mathbf{z}^{g_2})$ & & 69.1 \textcolor{red}{($\downarrow$1.5)} & 77.0 \textcolor{red}{($\downarrow$6.0)} & 84.5 \textcolor{red}{($\downarrow$2.1)} & 73.5/82.3\\
        \midrule
        $en(\mathbf{z}^{f_1}, \mathbf{z}^{g_1})$ & 19 & 67.5 \textcolor{red}{($\downarrow$3.1)} & 82.3 \textcolor{red}{($\downarrow$0.7)} & 86.9 \textcolor{ForestGreen}{($\uparrow$0.3)} & 76.6/83.8 \\
        \hline
        \midrule
        \multicolumn{6}{c}{Single-view with $f$ only: $a_{ij}=\cos(\mathbf{z}_i^f,\mathbf{z}_j^f)$} \\[0.5ex]
        \multicolumn{6}{c}{Single-view with $g$ only: $a_{ij}=\cos(\mathbf{z}_i^g,\mathbf{z}_j^g)$} \\[0.5ex]
        \midrule
        $\mathbf{z}^f$ & 9 & 57.8 \textcolor{red}{($\downarrow$9.6)} & 81.6 \textcolor{red}{($\downarrow$1.4)} & 85.8 \textcolor{red}{($\downarrow$0.8)} & 74.8/82.3 \\
        $\mathbf{z}^g$ & 1.5 & 68.7 \textcolor{red}{($\downarrow$0.5)} & 81.1 \textcolor{red}{($\downarrow$1.5)} & 83.3 \textcolor{red}{($\downarrow$1.9)} & 72.9/81.0 \\
        $en(\mathbf{z}^f, \mathbf{z}^g)$ & 10.5 & 68.6 \textcolor{red}{($\downarrow$2.0)} & 82.3 \textcolor{red}{($\downarrow$0.7)} & 86.3 \textcolor{red}{($\downarrow$0.3)} & 75.4/82.5 \\
        \hline
        \midrule
        \multicolumn{6}{c}{Multi-View with $f$ and $g$: $a_{ij}=\cos(\mathbf{z}_i^f,\mathbf{z}_j^f)+\cos(\mathbf{z}_i^g,\mathbf{z}_j^g)+\cos(\mathbf{z}_i^f,\mathbf{z}_j^g)+\cos(\mathbf{z}_i^g,\mathbf{z}_j^f)$} \\[0.5ex]
        \midrule
        $\mathbf{z}^f$ & \multirow{3}{*}{8} & 48.3 \textcolor{red}{($\downarrow$19.1)} & 79.9 \textcolor{red}{($\downarrow$3.1)} & 85.4 \textcolor{red}{($\downarrow$1.2)} & 74.9/83.1 \\
        $\mathbf{z}^g$ & & 68.8 \textcolor{red}{($\downarrow$0.4)} & 81.9 \textcolor{red}{($\downarrow$0.7)} & 84.0 \textcolor{red}{($\downarrow$1.2)} & 81.4/73.6 \\
        $en(\mathbf{z}^f, \mathbf{z}^g)$ & & 65.7 \textcolor{red}{($\downarrow$4.9)} & 82.3 \textcolor{red}{($\downarrow$0.7)} & 86.3 \textcolor{red}{($\downarrow$0.3)} & 75.9/83.3 \\
        \hline
        \midrule
        \multicolumn{6}{c}{Multi-View with $f$ and $g$: $a_{ij}=\cos(\mathbf{z}_i^f,\mathbf{z}_j^f)+\cos(\mathbf{z}_i^g,\mathbf{z}_j^g)$} \\[0.5ex]
        \midrule
        $\mathbf{z}^f$ & \multirow{3}{*}{8} & 59.9 \textcolor{red}{($\downarrow$7.5)} & 80.5 \textcolor{red}{($\downarrow$2.5)} & 85.2 \textcolor{red}{($\downarrow$1.4)} & 74.5/82.2 \\
        $\mathbf{z}^g$ & & 68.5 \textcolor{red}{($\downarrow$0.7)} & 80.2 \textcolor{red}{($\downarrow$2.4)} & 83.0 \textcolor{red}{($\downarrow$2.2)} & 68.5/80.7 \\
        $en(\mathbf{z}^f, \mathbf{z}^g)$ & & 67.5 \textcolor{red}{($\downarrow$3.1)} & 82.0 \textcolor{red}{($\downarrow$1.0)} & 86.2 \textcolor{red}{($\downarrow$0.4)} & 75.0/82.4 \\
        \hline
        \midrule
    \end{tabular}
\end{center}
\label{UMBCmvsv}
\end{table}

\textbf{In general, an ensemble of the representations generated from two distinct encoding functions performs even better.} The two encoding functions, $f$ and $g$, have naturally different behaviour. Although the $f$ function, which is an RNN, is able to approximate a linear function $g$, the distributional similarity \cite{firth57synopsis}, which implies that spatially adjacent sentences should be mapped to close vectors\footnote{Without the distributional similarity, the model learns a trivial solution that matches the representations produced from $f$ and $g$ for the same input sentence, since $f$ is  powerful and  able to approximate $g$. In this case, both $f$ and $g$ collapse, and no useful feature is learnt after training.}, helps the two functions to learn more generalised representations. Therefore, $f$ and $g$ encode the input sentence with emphasis on different aspects, and the linear model that is subsequently trained for each of the supervised downstream tasks benefits from this diversity leading to better predictions.

Compared with the ensemble of two multi-view models, each with two encoding functions of the same type, our multi-view model with $f$ and $g$ provides slightly better results on STS tasks, and similar results on supervised evaluation tasks, while our model has much higher training efficiency. Compared with the ensemble of two single-view models, each with only one encoding function, the matching between $f$ and $g$ in our multi-view model produces better results. 

\subsection{Symmetric Agreement Between Two Views}
\begin{align}
    a_{ij}&=[\hat{\mathbf{z}}_i^f;\hat{\mathbf{z}}_j^g]^\top[\hat{\mathbf{z}}_i^g;\hat{\mathbf{z}}_j^f]=\cos(\mathbf{z}_i^f,\mathbf{z}_j^g)+\cos(\mathbf{z}_i^g,\mathbf{z}_j^f) \label{agm1} \\
    a_{ij}&=(\hat{\mathbf{z}}_i^f+\hat{\mathbf{z}}_j^g)^\top(\hat{\mathbf{z}}_i^f+\hat{\mathbf{z}}_j^g)=\cos(\mathbf{z}_i^f,\mathbf{z}_j^f)+\cos(\mathbf{z}_i^g,\mathbf{z}_j^g)+\cos(\mathbf{z}_i^f,\mathbf{z}_j^g)+\cos(\mathbf{z}_i^g,\mathbf{z}_j^f) \label{agm2} \\
    a_{ij}&=[\hat{\mathbf{z}}_i^f;\hat{\mathbf{z}}_j^g]^\top[\hat{\mathbf{z}}_i^f;\hat{\mathbf{z}}_j^g]=\cos(\mathbf{z}_i^f,\mathbf{z}_j^f)+\cos(\mathbf{z}_i^g,\mathbf{z}_j^g) \label{agm3}
\end{align}
Many choices are plausible for calculating the agreement between the two distinctive views in our proposed learning framework, thus it is important to empirically compare a few reasonable ones.
Besides the one we used in our learning framework, two other symmetric agreement functions were tested. The definition of the three agreement functions are listed as Eq. \ref{agm1}, \ref{agm2}, and \ref{agm3}, where $\hat{\mathbf{z}}$ denotes the post-processed and normalised vector, $\mathbf{z}$ denotes the post-processed vector. The results are presented in the last two sections in Table \ref{UMBCmvsv}.

\begin{figure}
    \centering
    \begin{subfigure}[b]{0.3\textwidth}
        \includegraphics[width=\textwidth]{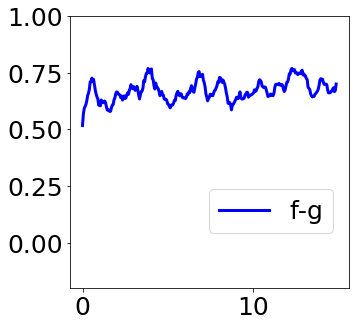}
        \caption{Trained with \textbf{our} Eq. \ref{agm1}.}
        \label{fgaddfg}
    \end{subfigure}
    ~
    \begin{subfigure}[b]{0.3\textwidth}
        \includegraphics[width=\textwidth]{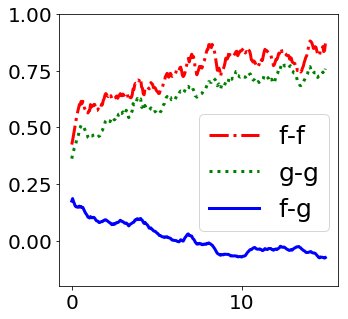}
        \caption{Trained with Eq. \ref{agm2}.}
        \label{faddgfaddg}
    \end{subfigure}
    \caption{\textbf{Mean cosine similarity of adjacent sentences divided by temperature vs. number of training iterations/10k}. ``$f$-$f$'' refers to $\cos(\mathbf{z}_i^f,\mathbf{z}_j^f)/\tau$. As we can see, during training with Eq. \ref{agm2}, $f$-$f$ and $g$-$g$ gradually dominate the agreement, thus $f$-$g$ gets down-weighted. While with our Eq. \ref{agm1}, $f$ and $g$ are constantly encouraged to learn from the other.}
    \label{cosine}
\end{figure}

We found that the model with the agreement in Eq.\ref{agm1}, which is used in all of our other experiments, outperforms those with the other two agreement functions. Our explanation is that, in both Eq.\ref{agm2} and Eq.\ref{agm3}, maximising the agreement among the representations from one single view is involved, and since the representations produced from the same function, either $f$ or $g$, tend to have a similar structure, it is easier to optimise each of the two views to match itself (on neighbouring sentences), instead of the other one, which conflicts with the goal of multi-view learning (see Figure \ref{cosine}). 

\section{Conclusion} 
We proposed a unified multi-view sentence representation learning framework that combines an RNN-based encoder and an average-on-word-vectors linear encoder and can be efficiently trained within a few hours on a large unlabelled corpus. The experiments were conducted on three large unlabelled corpora, and meaningful comparisons were made to demonstrate the generalisation ability and the transferability of our learning framework, and also to consolidate our claim. The produced sentence representations outperform existing unsupervised transfer methods on unsupervised evaluation tasks, and match the performance of the best unsupervised model on supervised evaluation tasks. 

As presented in our experiments, the ensemble of two views leveraged the advantages of both views, and provides rich semantic information of the input sentence, also the multi-view training helps each view to produce better representations than the single-view training does. Meanwhile, our experimental results also support the finding in \cite{Hill2016LearningDR} that linear/log-linear models ($g$ in our model) tend to work better on the unsupervised tasks, while RNN-based models ($f$ in our model) generally perform better on the supervised tasks. Future work should explore the relaxation of the cosine similarity metric to incorporate the length information of the produced sentence representations.

Our multi-view learning framework was inspired by the asymmetric information processing in the two hemispheres of the human brain, in which for most adults, the left hemisphere contributes to sequential processing, including primarily language understanding, and the right one carries out more parallel processing, including visual spatial understanding \cite{bryden2012laterality}. The experimental results raise an intriguing hypothesis about how these two types of information processing may complementarily help learning.

\section*{Acknowledgements}
We appreciate the gift funding from Adobe Research. Many thanks to Sam Bowman for the thoughtful discussion, and to Mengting Wan, Wangcheng Kang, and Jianmo Ni for critical comments on the project.




\bibliography{nips_2018}

\clearpage

\begin{center}
    \LARGE{\textbf{Supplementary Material}}
\end{center}

\setcounter{section}{0}
\setcounter{table}{0}

\section{Power Iteration}
The Power Iteration was proposed in \cite{mises1929praktische}, and it is an efficient algorithm for estimating the top eigenvector of a given covariance matrix. Here, it is used to estimate the top principal component from the representations produced from $f$ and $g$ separately. We omit the superscription here, since the same step is applied to both $f$ and $g$.

Suppose there is a batch of representations $\mathbf{Z}=[\mathbf{z}_1,\mathbf{z}_2\,...,\mathbf{z}_N]\in\mathbb{R}^{2d\times N}$ from either $f$ or $g$, 
the Power Iteration method is applied here to estimate the top eigenvector of the covariance matrix\footnote{In practice, usually $N$ is less than $2d$, thus we estimate the top eigenvector of $\mathbf{Z}^\top\mathbf{Z} \in \mathbb{R}^{N\times N}$.}: $\mathbf{C}=\mathbf{Z}\mathbf{Z}^\top$, and it is described in Algorithm \ref{A1}:

\begin{algorithm}[H]
\caption{Estimating the First Principal Component (\textbf{Power Iteration} \cite{mises1929praktische})}
\begin{algorithmic}[1]
\Statex \textbf{Input:} Covariance matrix $\mathbf{C}\in\mathbb{R}^{2d\times2d}$, number of iterations $T$
\Statex \textbf{Output:} First principal component $\mathbf{u}\in\mathbb{R}^{2d}$
\State Initialise a unit length vector $\mathbf{u}\in\mathbb{R}^{2d}$
\For{$t\gets 1$, $T$}
\State $\mathbf{u}\gets \mathbf{C}\mathbf{u}$, 
\State $\mathbf{u}\gets \displaystyle\frac{\mathbf{u}}{||\mathbf{u}||}$
\EndFor
\end{algorithmic}
\label{A1}
\end{algorithm}

\section{Detailed Results on STS tasks}

Every year, STS task has multiple datasets, so detailed comparison on every dataset is helpful to understand the behaviour of our model and the related work. We present the detailed results on all datasets in Table \ref{detailed_comp}, and since FastSent \cite{Hill2016LearningDR} and QT \cite{logeswaran2018an} only reported their performance on STS14, we use a separate Table \ref{fsqt_details} to compare their models with ours.

We tuned the hyperparameters in our model trained on the BookCorpus \cite{Zhu2015AligningBA} on the averaged Pearson's score on STS14, and it is clear that our model performs better than others on STS14 on average. Although there are some dataset overlaps among STS12, 13, 14, 15 and 16, our model still works better other models on those datasets that are not in overlap with STS14, which means that our model demonstrates a solid generalisation ability and transferability.

\begin{table}[t]
\fontsize{8.5}{11}\selectfont
\begin{adjustwidth}{-0.7cm}{}
\caption{\textbf{Results on unsupervised evaluation tasks} (Pearson's $r\times 100$). \textbf{Bold} numbers are the best results among unsupervised transfer models, and \underline{underlined} numbers are the best ones among all models.}
\begin{tabular}{| c || K{0.45cm} | K{0.4cm} | K{0.4cm} | K{0.3cm} K{0.3cm} K{0.3cm} K{0.5cm} | K{0.3cm} K{0.4cm} | K{0.4cm} | K{0.5cm} || K{0.6cm} | K{0.7cm} | K{0.7cm} |}
\hline
\multirow{3}{*}{Dataset} & \multicolumn{11}{c||}{Un. Transfer \cite{Arora2017ASB,Mu2017AllbuttheTopSA,Kenter2016SiameseCO}} & \multicolumn{3}{c|}{Su. Transfer \cite{Conneau2017SupervisedLO,Wieting2017RevisitingRN}} \\
\cline{2-15}
 & \multicolumn{3}{c|}{\textbf{Multi-view}} & \multicolumn{4}{c|}{GloVe} & \multicolumn{2}{c|}{word2vec} & \multirow{2}{*}{ST}  & S- & Infer & \multirow{2}{*}{GRAN} & LSTM \\
\cline{2-10}
 & C1 & C2 & C3 & avg & tfidf & WR & proc. & bow & proc. &  & cbow & Sent &  & avg \\
 \hline
MSRpar & 40.3 & 43.0 & 40.1 & \textbf{47.7} &  50.3  & 35.6 & 44.1 & 42.1 & 43.9 & 16.8  & 43.8 & 40.0 & 47.7 & \underline{49.0} \\
MSRvid & 85.4 & \underline{\textbf{87.8}} & 84.8 & 63.9 & 77.9 &   83.8 & 68.1 & 72.1 & 72.2 & 41.7 & 45.2 & 82.8 &   85.2  & 84.3 \\
SMTeuro & 51.2 & 54.2 & 51.1 & 46.0 &  54.7  & 49.9 & 45.3 & 53.2 & \underline{\textbf{54.3}} & 35.2  & 45.0 & 49.6 & 49.3 & 51.2 \\
OnWN & 74.2 & \underline{\textbf{74.8}} & 72.8 & 55.1 & 64.7 & 66.2 & 65.7 & 69.4 & 69.5 & 29.7  & 64.4 & 59.6 & 71.5 & 71.5 \\
SMTnews & 53.3 & \textbf{60.3} & 54.5 &   59.6  & 45.7 & 45.6 & 47.2 & 49.4 & 48.5 & 30.8  & 39.0 & 59.3 & 58.7 &   \underline{68.0}  \\
\hline
STS'12 & 60.9 & \textbf{64.0} & 46.8 & 52.5 & 58.7 & 56.2 & 54.1 & 57.2 & 57.7 & 30.8  & 47.5 & 58.2 & 62.5 &   \underline{64.8}   \\
\hline
FNWN & 46.3 & \textbf{47.9} & 46.8 & 34.2 & 36.6 & 39.4 & 39.3 & 40.7 & 42.0 & 30.4  & 23.2 & 26.3 &   \underline{55.6}  & 53.2 \\
headlines & 69.9 & \textbf{74.4} & 73.4 & 63.8 & 63.7 & 64.7 & 57.2 & 61.9 & 63.8 & 34.6 & 65.3 & 66.4 & 76.1 &   \underline{77.3}  \\
OnWN & \underline{\textbf{83.4}} & 82.9 & 80.2 & 49.0 & 75.2 & 82.8 & 58.6 & 67.9 & 68.2 & 10.0 & 49.9 & 69.2 &   81.4  & 81.2 \\
SMT & 40.8 & \underline{\textbf{41.5}} & 39.3 & 22.3 & 29.6 & 37.9 & - & - & - & 24.3 &  33.1 & 32.0 & 40.3 &   40.7  \\
\hline
STS'13 & 60.1 & \textbf{61.7} & 59.9 & 42.3 & 52.1 & 56.6 & 57.7 & 56.8 & 58.0 & 24.8  & 42.9 & 48.5 &   \underline{63.4}  & 63.1 \\
\hline
deft-forum & \textbf{51.0} & \textbf{51.0} & 44.3 & 27.1 & 37.5 & 41.2 & 29.4 & 32.2 & 33.3 & 12.9  & 40.8 & 42.4 &   \underline{55.7}  & 56.6 \\
deft-news & 67.6 & \textbf{73.3} & 72.0 & 68.0 & 68.7 &   69.4  & 71.5 & 66.8 & 66.0 & 23.5  & 59.1 & 73.3 & 77.1 &   \underline{78.0}  \\
headlines & 66.8 & \textbf{71.8} & 68.4 & 59.5 & 63.7 & 64.7 & 52.6 & 58.0 & 59.6 & 37.8  & 63.6 & 61.7 & 72.8 &   \underline{74.5}  \\
images & 83.1 & \underline{\textbf{86.2}} & 84.1 & 61.0 & 72.5 & 82.6 & 68.3 & 73.8 & 74.2 & 51.2  & 65.0 & 78.5 &   85.8  & 84.7 \\
OnWN & \textbf{84.2} & 84.1 & 81.7 & 58.4 & 75.2 &   82.8  & 67.6 & 74.6 & 74.8 & 23.3  & 60.7 & 76.5 &   \underline{85.1}  & 84.9 \\
tweet-news & \textbf{76.1} & 75.8 & 73.4 & 51.2 & 65.1 & 70.1 & 66.1 & 71.9 & 72.1 & 39.9  & 75.2 & 70.0 &   \underline{78.7}  & 76.3 \\
\hline
STS'14 & 71.5$\star$ & \textbf{73.4} & 70.7 & 54.2 & 63.8 & 68.5 & 59.2 & 62.9 & 63.3 & 31.4  & 60.4 & 67.1 &   \underline{75.9}  & 75.8 \\
\hline
answers-forums & 72.0 & 72.6 & \textbf{72.7} & 30.5 & 45.6 & 63.9 & 39.9 & 46.4 & 46.8 & 36.1 & 21.8 & 60.5 &   \underline{73.1}  & 71.8 \\
answers-students & 74.3 & 71.0 & \underline{\textbf{74.7}} & 63.0 & 63.9 & 70.4 & 62.4 & 68.1 & 68.0 & 33.0 & 36.7 & 68.0 & 72.9 & 71.1 \\
belief & \underline{\textbf{79.0}} & 77.9 & 75.9 & 40.5 & 49.5 & 71.8 & 57.7 & 59.7 & 60.4 & 24.6 & 47.7 & 71.5 &   78.0  & 75.3 \\
headlines & 72.7 & \textbf{77.9} & 75.7 & 61.8 & 70.9 & 70.7 & 53.3 & 61.5 & 63.5 & 43.6 & 21.5 & 70.4 & 78.6 &   \underline{79.5}  \\
images & 84.3 & \underline{\textbf{86.4}} & 83.8 & 67.5 & 72.9 & 81.5 & 73.2 & 78.1 & 78.1 & 17.7 & 25.6 & 85.0 &   85.8  &   85.8  \\
\hline
STS'15 & 76.4 & \textbf{77.2} & 76.5 & 52.7 & 60.6 & 71.7 & 57.3 & 62.7 & 63.4 & 31.0  & 30.7 & 71.1 &   \underline{77.9}  &  76.7 \\
\hline
answer-answer & \textbf{68.7} & 65.1 & 64.3 & - & - & - & - & - & - & - & -  & 61.1 & - & -  \\
headlines & 71.7 & \textbf{75.0} & 73.4 & - & - & - & - & - & - & - & - & 68.6 & - & -    \\
plagiarism & 84.4 & \textbf{84.8} & 83.7 & - & - & - & - & - & - & - & -  & 80.5 & - & -  \\
postediting & 85.3 & 84.3 & \textbf{85.9} & - & - & - & - & - & - & - & - & 81.9 & - & - \\
question-question & 68.9 & \textbf{74.1} & 66.4 & - & - & - & - & - & - & - & -  & 64.0 & - & - \\
\hline
STS'16 & 75.8 & \textbf{76.7} & 74.8 & - & - & - & - & - & - & - & -  & 71.2 & - & - \\
\hline
SICK'14 & 74.7 & \underline{\textbf{74.9}} & 72.8 & 69.4 & 69.4 & 72.2& 67.9 & 70.1 & 61.5 & 49.8 & -  & 73.4 & 72.9 & 71.3  \\
\hline
\end{tabular}
\label{detailed_comp}
\end{adjustwidth}
\end{table}

\begin{table}[t]
\caption{Comparison with FastSent and QT on every datasets in STS14 (Pearson's $r\times 100$).}
\begin{center}
\begin{tabular}{|c|c|c|c|c|c c c|}
\hline
\multirow{2}{*}{Datasets}& \multicolumn{2}{c}{FastSent \cite{Hill2016LearningDR}} & \multicolumn{2}{|c|}{QT \cite{logeswaran2018an}} & \multicolumn{3}{c|}{\textbf{Multi-view}} \\
\cline{2-8}
& & +AE & RNN & BOW & C1 & C2 & C3 \\
\hline
deft-forum & 41.0 & 41.0 & 15.0 & 37.0 & \textbf{51.0} & \textbf{51.0} & 44.3 \\
deft-news & 58.0 & 56.0 & 48.0 & 62.0 & 67.6 & \textbf{73.3} & 72.0 \\
deft-headlines & 57.0 & 58.0 & 48.0 & 60.0 & 66.8 & \textbf{71.8} & 68.4 \\
images & 74.0 & 63.0 & 53.0 & 76.0 & 83.1 & \textbf{86.2} & 84.1 \\
OnWN & 74.7 & 69.0 & 53.0 & 76.0 & \textbf{84.2} & 84.1 & 81.7 \\
tweet-news & 63.0 & 70.0 & 62.0 & 67.0 & \textbf{76.1} & 75.8 & 73.4 \\
\hline
STS14 & 61.2 & 59.5 & 49.0 & 65.0 & 71.5$\star$ & \textbf{73.4} & 70.7 \\
\hline
\end{tabular}
\end{center}
\label{fsqt_details}
\end{table}

\begin{table}[t]
\caption{Results of our multi-view model with $f$ and $g$ and other variants. In the table, ``Avg of STS tasks'' refers to the mean Pearson's score on 5 STS tasks; ``Avg of SICK-R, STS-B'' refers to the mean Pearson's score on Sick-Entailment and STS-Benchmark as they both require the same feature engineering methods proposed in \cite{Tai2015ImprovedSR}; ``Avg of Binary-CLS tasks'' refers to the mean accuracy on 5 sentiment analysis tasks; ``MRPC'' refers to the Microsoft Paraphrase Detection task, and the results are reported in Accuracy/F1-score. $en(\cdot,\cdot)$ stands for an ensemble of 2 representations. Better view in colour.}
\fontsize{8.5}{11}\selectfont
\begin{adjustwidth}{-0cm}{}
\begin{center}
    \begin{tabular}{ c | c | c | c | c | c }
    \hline 
        \midrule
        \textbf{BookCorpus} & \multirow{2}{*}{Hrs} & Avg of STS tasks  & Avg of & Avg of Binary-CLS tasks & \multirow{2}{*}{MRPC} \\[0.5ex]
        (C1) & & (STS12-16) & SICK-R, STS-B & (MR, CR, SUBJ, MPQA, SST) &  \\[0.5ex]
        \Xhline{2pt}
        \midrule
        \multicolumn{6}{c}{Multi-View with $f$ and $g$: $a_{ij}=\cos(\mathbf{z}_i^f,\mathbf{z}_j^g)+\cos(\mathbf{z}_i^g,\mathbf{z}_j^f)$} \\[1ex]
        \midrule
        $\mathbf{z}^f$ & \multirow{3}{*}{3} & 64.4 & 81.3 & 86.5 & 75.1/82.7\\
        $\mathbf{z}^g$ & & 68.7 & 82.6 & 85.2 & 74.0/81.5 \\
        $en(\mathbf{z}^f, \mathbf{z}^g)$ & & 68.9 & 82.6 & 87.0 & 77.1/83.4 \\
        \Xhline{2pt}
        \midrule
        \multicolumn{6}{c}{Multi-view with $f_1$ and $f_2$: $a_{ij}=\cos(\mathbf{z}_i^{f_1},\mathbf{z}_j^{f_2})+\cos(\mathbf{z}_i^{f_2},\mathbf{z}_j^{f_1})$} \\[1ex]
        \multicolumn{6}{c}{Multi-view with $g_1$ and $g_2$: $a_{ij}=\cos(\mathbf{z}_i^{g_1},\mathbf{z}_j^{g_2})+\cos(\mathbf{z}_i^{g_2},\mathbf{z}_j^{g_1})$} \\[1ex]
        \midrule
        $\mathbf{z}^{f_1}$ & \multirow{2}{*}{6} & 62.3 \textcolor{red}{($\downarrow$2.1)} & 81.6 \textcolor{ForestGreen}{($\uparrow$0.3)} & 86.3 \textcolor{red}{($\downarrow$0.2)} & 75.7/83.7  \\
        $en(\mathbf{z}^{f_1}, \mathbf{z}^{f_2})$ & & 63.1 \textcolor{red}{($\downarrow$5.8)} & 81.9 \textcolor{red}{($\downarrow$0.7)} & 87.1 \textcolor{ForestGreen}{($\uparrow$0.1)} & 76.7/83.3 \\
        \midrule
        $\mathbf{z}^{g_1}$ & \multirow{2}{*}{1} & 68.2 \textcolor{red}{($\downarrow$0.5)} & 81.8 \textcolor{red}{($\downarrow$0.8)} & 85.0 \textcolor{red}{($\downarrow$0.2)} & 73.7/81.7 \\
        $en(\mathbf{z}^{g_1}, \mathbf{z}^{g_2})$ & & 69.0 \textcolor{ForestGreen}{($\uparrow$0.1)} & 78.0 \textcolor{red}{($\downarrow$4.6)} & 85.3 \textcolor{red}{($\downarrow$1.7)} & 73.8/82.6\\
        \midrule
        $en(\mathbf{z}^{f_1}, \mathbf{z}^{g_1})$ & 7 & 68.6 \textcolor{red}{($\downarrow$0.3)} & 82.6 \textcolor{RoyalBlue}{($-$)} & 87.0 \textcolor{RoyalBlue}{($-$)} & 76.5/84.2 \\
        \hline
        \midrule
        \multicolumn{6}{c}{Single-view with $f$ only: $a_{ij}=\cos(\mathbf{z}_i^f,\mathbf{z}_j^f)$} \\[1ex]
        \multicolumn{6}{c}{Single-view with $g$ only: $a_{ij}=\cos(\mathbf{z}_i^g,\mathbf{z}_j^g)$} \\[1ex]
        \midrule
        $\mathbf{z}^f$ & 2.5 & 60.9 \textcolor{red}{($\downarrow$3.5)} & 81.5 \textcolor{ForestGreen}{($\uparrow$0.2)} & 86.2 \textcolor{red}{($\downarrow$0.3)} & 74.7/82.6 \\
        $\mathbf{z}^g$ & 1 & 68.7 \textcolor{RoyalBlue}{($-$)} & 82.3 \textcolor{red}{($\downarrow$0.3)} & 85.1 \textcolor{red}{($\downarrow$0.1)} & 72.8/81.8 \\
        $en(\mathbf{z}^f, \mathbf{z}^g)$ & 3.5 & 67.2 \textcolor{red}{($\downarrow$1.7)} & 82.7 \textcolor{ForestGreen}{($\uparrow$0.1)} & 87.1 \textcolor{ForestGreen}{($\uparrow$0.1)} & 76.4/83.2 \\
        \hline
        \midrule
        \multicolumn{6}{c}{Multi-View with $f$ and $g$: $a_{ij}=\cos(\mathbf{z}_i^f,\mathbf{z}_j^f)+\cos(\mathbf{z}_i^g,\mathbf{z}_j^g)+\cos(\mathbf{z}_i^f,\mathbf{z}_j^g)+\cos(\mathbf{z}_i^g,\mathbf{z}_j^f)$} \\[1ex]
        \midrule
        $\mathbf{z}^f$ & \multirow{3}{*}{3} & 61.3 \textcolor{red}{($\downarrow$3.1)} & 80.6 \textcolor{red}{($\downarrow$0.7)} & 86.5 \textcolor{RoyalBlue}{($-$)} & 74.8/81.9 \\
        $\mathbf{z}^g$ & & 68.2 \textcolor{red}{($\downarrow$0.5)} & 82.6 \textcolor{RoyalBlue}{($-$)} & 84.8 \textcolor{red}{($\downarrow$0.4)} & 74.7/82.2 \\
        $en(\mathbf{z}^f, \mathbf{z}^g)$ & & 64.5 \textcolor{red}{($\downarrow$4.4)} & 82.6 \textcolor{RoyalBlue}{($-$)} & 86.9 \textcolor{red}{($\downarrow$0.1)} & 75.8/83.0 \\
        \hline
        \midrule
        \multicolumn{6}{c}{Multi-View with $f$ and $g$: $a_{ij}=\cos(\mathbf{z}_i^f,\mathbf{z}_j^f)+\cos(\mathbf{z}_i^g,\mathbf{z}_j^g)$} \\[1ex]
        \midrule
        $\mathbf{z}^f$ & \multirow{3}{*}{3} & 52.9 \textcolor{red}{($\downarrow$11.5)} & 77.7 \textcolor{red}{($\downarrow$3.6)} & 85.2 \textcolor{red}{($\downarrow$1.3)} & 74.7/82.2 \\
        $\mathbf{z}^g$ & & 67.8 \textcolor{red}{($\downarrow$0.9)} & 81.8 \textcolor{red}{($\downarrow$0.8)} & 84.1 \textcolor{red}{($\downarrow$1.1)} & 72.7/81.9 \\
        $en(\mathbf{z}^f, \mathbf{z}^g)$ & & 64.6 \textcolor{red}{($\downarrow$4.3)} & 82.6 \textcolor{RoyalBlue}{($-$)} & 87.0 \textcolor{RoyalBlue}{($-$)} & 77.1/83.4 \\
        \hline
        \midrule
    \end{tabular}
\end{center}
\label{BCmvsv}
\end{adjustwidth}
\end{table}

\section{Multi-view Learning vs. Single-view Learning}
In order to show that the multi-view learning with $f$ and $g$ is helping the learning, we compare our model with other variants, including the multi-view model with 2 functions of the same type but parameterised independently, either 2 $f$-s or 2 $g$-s, and the single-view model with only one $f$ or $g$. The results of models trained on BookCorpus with different settings are presented in in Table \ref{BCmvsv}. 

The results also support our claim that our multi-view learning with 2 different views improves each view in single-view learning, and also performs better than the multi-view models with the same architecture but parameterised separately. 

Generally, ensemble produces better results on supervised evaluation tasks. However, only in our multi-view learning framework with 2 distinctive encoders, an ensemble of 2 representations provides better performance on STS tasks. The performance of an ensemble of 2 representations in other variants is inferior to that of the linear encoding function $g$ itself. 


\section{Training \& Model Details}
The hyperparameters we need to tune include the batch size $N$, the dimension of the GRU encoder $d$, and the context window $c$, and the results we presented in this paper is based on the model trained with $N=512$, $d=1024$, and $c=3$. It takes up to 8GB on a GTX 1080Ti GPU. 

The initial learning rate is $5\times 10^{-4}$, and we didn't anneal the learning rate through the training. All weights in the model are initialised using the method proposed in \cite{He2015DelvingDI}, and all gates in the bi-GRU are initialised to 1, and all biases in the single-layer neural network are zeroed before training. The word vectors are fixed to be those in the FastText \cite{Bojanowski2017EnrichingWV}, and we don't finetune them. Words that are not in the FastText's vocabulary are fixed to $0$ vectors through training. The temperature term is initialised as $1$, and is tuned by the gradient descent during training.

The temperature term is used to convert the agreement $a_{ij}$ to a probability distribution $p_{ij}$ in Eq. 1 in the main paper. In our experiments, $\tau$ is a trainable parameter initialised to $1$ that decreased consistently through training. Another model trained with fixed $\tau$ set to the final value performed similarly.


\section{Number of Parameters}
The number of parameters of each of the selected models is:
\begin{enumerate}
\item Ours: $6\times d \times d \times 2 + 300 \times 2d \approx 13.2M$
\item Quick-thought \cite{logeswaran2018an}: $\approx 19.8M$ 
\item Skip-thought \cite{Kiros2015SkipThoughtV}: $\approx 57.7M$ 
\end{enumerate}

\end{document}